\pdfoutput=1
\documentclass[11pt]{article}

\usepackage{acl}

\usepackage{times}
\usepackage{latexsym}

\usepackage[T1]{fontenc}
\usepackage[utf8]{inputenc}

\usepackage{microtype}
\usepackage{graphicx}
\usepackage{amsmath}
\usepackage{array}
\usepackage{lipsum}
\usepackage{lmodern}
\usepackage{tcolorbox}
\usepackage{multicol}
\usepackage{booktabs}
\usepackage{verbatim}

\newcolumntype{L}[1]{>{\raggedright\let\newline\\\arraybackslash\hspace{0pt}}m{#1}}
\newcolumntype{C}[1]{>{\centering\let\newline\\\arraybackslash\hspace{0pt}}m{#1}}
\newcolumntype{R}[1]{>{\raggedleft\let\newline\\\arraybackslash\hspace{0pt}}m{#1}}

\usepackage{placeins}
\usepackage{float}
\usepackage{xcolor}

\title{Disentangling Knowledge-based and Visual Reasoning by Question Decomposition in KB-VQA}

\author{Elham J. Barezi \\
  Michigan State University \\
  \texttt{jebalbar@msu.edu} \\\And
  Parisa Kordjamshidi \\
  Michigan State University \\
  \texttt{kordjams@msu.edu} \\}

\begin{document}
\maketitle

\begin{abstract}
We study the Knowledge-Based visual question-answering problem, for which given a question, the models need to ground it into the visual modality to find the answer. Although many recent works use question-dependent captioners to verbalize the given image and use Large Language Models to solve the VQA problem, the research results show they are not reasonably performing for multi-hop questions. 
%We hypothesize that the captioners are not suitable for complex questions since they have been merely trained to verbalize the salient objects, and often miss the details required for multi-hop reasoning. In particular, in the KB-VQA case, multiple hops require different sources of information that are not necessarily contained in the visual modality.

%To deal with this challenge our contribution has multiple facets. We follow the approach of question decomposition in the first place. 
Our study shows that replacing a complex question with several simpler questions helps to extract more relevant information from the image and provide a stronger comprehension of it. Moreover, we analyze the decomposed questions to find out the modality of the information that is required to answer them and use a captioner for the visual questions and LLMs as a general knowledge source for the non-visual KB-based questions.
Our results demonstrate the positive impact of using simple questions before retrieving visual or non-visual information. We have provided results and analysis on three well-known VQA datasets including OKVQA, A-OKVQA, and KRVQA, and achieved up to 2\% improvement in accuracy.
\end{abstract}

\section{Introduction}

Knowledge Base Visual Question Answering (KB-VQA)  aims to answer a question in natural language form while visual and external information is needed to provide the correct answer. Though over the past years, the progress in this field has been remarkable, existing models still suffer in verbalizing the images and processing long and complicated questions. 
  
With the introduction of the pre-trained large-scale language models and their vast power in solving various tasks, recent approaches aimed to use the implicit knowledge accumulated in these models, such as GPT-3~\cite{brown2020language}. These models \cite{hu2023promptcap, shao2023prompting,yang2022empirical} tried to use LLMs and guide commonsense reasoning in VQA models. 

Despite their remarkable results, these methods lack strong visual comprehension. They verbalize the image using a general or question-dependent caption, while the captioners are not suitable for complex questions since they have been merely trained to verbalize the salient objects, and often miss the details required for multi-hop reasoning. In particular, in the KB-VQA case, multiple hops require different sources of information that are not necessarily contained in the visual modality. The captioners are not strong in analyzing multi-hop questions and stick to the main objects in the image rather than focusing on the required details. Generating a caption using only one question can generate one explanation for the image, though asking multiple relevant questions helps uncover more relevant required visual details and provide a richer verbalization of the image. Moreover, in the KB-VQA case, multiple hops may require different sources of information that are not necessarily contained in the visual modality, and decomposing the questions helps to find non-visual parts of the question to retrieve the extra required information.

In this work,  we propose a question decomposer to find several simpler questions to guide the captioner and provide a richer textual representation of the given image. In addition, we decide whether the decomposed questions are visual or non-visual. We use models such as PromptCap \cite{hu2023promptcap} or InstructBlip \cite{dai2024instructblip} for the visual questions to extract the required visual information. In addition, we use GPT models for the non-visual questions to extract extra knowledge required to answer the question. Finally, we use all the extracted information to help LLMs to find the final answer. 

 In a nutshell, our contributions are as follows: 
\begin{itemize}
    \itemsep0em 
    \item We address some weaknesses of current image-to-text captioners for KB-VQA problems, including question decomposition to extract more visual details required to address the given question.
    \item We perform type-checking for the simplified questions to cover both visual and knowledge-based information. 
    \item We show that enriching information by prompting the captioners/knowledge-base using various questions improves the final accuracy for the KB-VQA task.
\end{itemize}

\section{Related Works}
We roughly divide recent KB-VQA approaches into three groups as follows. The first group, plain VQA methods, focuses on question and image integration approaches without including any extra information.  The second group aims to feed extra information extracted from external knowledge bases in VQA models. In the last group of SOTA KB-VQA models, we review methods that rely on the power of the pre-trained Large Language Models (LLMs) as implicit sources of knowledge for KB-VQA tasks. 

\subsection{Plain VQA methods}
Both \cite{yu2018beyond} and MUTAN \cite{ben2017mutan} aim to use bilinear models for merging visual and textual information in VQA tasks using a multimodal Tucker tensor decomposition \cite{tucker1966some}. MUTAN parametrizes bilinear interactions between visual and textual representations to learn high-level associations between question meaning and visual concepts in the image while avoiding the curse of dimensionality. 

To focus on both the visual and textual context of questions, the authors of MCAN \cite{yu2019deep} aim to design a ‘co-attention’ model to associate keywords in questions with key objects in images. 
They propose a deep Modular Co-Attention Network (MCAN) made by in-depth cascading of Modular Co-Attention (MCA) layers. Each  Modular Co-Attention layer (MCA layer) models the self-attention of questions and images, as well as the question-guided-attention of images jointly using a modular composition of two basic attention units.
The authors of~\cite{ding2022mukea} suggest extracting knowledge triplets from training data, instead of feeding external knowledge bases. 

\subsection{KB processing methods}
%\pk{The related work seems like you focus on a few works and explain them to an extent without providing a big picture. Why MuCKO is important to start with? why it is related to your approach? Maybe mention that "Using the captioners for VQA, was first used in XXX...." We need to see a story in the related work not a list of selected papers with explanations in front of them without a clear motivation of why those are selected and how they are relevant to your approach. I hope this helps you to improve this section. The same applies to the above section too.} 
The MuCKO model~\cite{zhu2020mucko} aims to answer the question by generating a semantic graph built on the dense captions of the image, a graph representing relevant knowledge triplets, and a spatial visual graph where nodes represent region features. It performs question-guided inter and intra-graph attention to answer the question iteratively. 

\citeauthor{marino2021krisp} use Multimodal-BERT (MMBERT) \cite{khare2021mmbert} for their multimodal embedding and use Relational Graph Convolutional Network (RGCN) \cite{schlichtkrull2018modeling} to learn the representation of the concepts extracted from questions, Image, and KBs. 
MAVEx is proposed by~\cite{wu2022multi} aiming to leverage and integrate knowledge from visual (Google image search), textual (Wikipedia), and commonsense knowledge sources (ConceptNet). Their goal is to validate a set of candidate answers using answer-based knowledge retrieval.  

\citeauthor{guo2022unified} offer Unifer, a unified end-to-end retriever-reader framework for KB-VQA. To reduce noisy retrieval from external KBs, they evaluate retrieved triplets based on their effect on final accuracy. They generate pseudo-labels to evaluate all knowledge triplets for each question which is computationally very costly. 

DMMGR \cite{li2022dynamic} represents the retrieved KB triplets in dynamic key-value memory format and converts visual content into a spatial graph representation. % whose nodes are the object category embeddings, and edges are the relative position embeddings of two objects.  
It performs a question-knowledge-guided attention over the spatial visual graph to find the required visual information to answer the question.

\subsection{Large Language Models for VQA task}

Due to the revolutionary growth and power of large pre-trained models, the new trend in knowledge-based VQA aims to use large language-only pre-trained models like GPT-3 \cite{brown2020language} as implicit sources of knowledge, instead of using external sources of knowledge. Most often, these methods replace the visual data (image) with its corresponding tags and/or captions and continue to solve an unimodal text-only problem. 

\citeauthor{gui2021kat} in KAT use a contrastive-learning-based module to retrieve knowledge from an external KB and use GPT-3 to retrieve implicit knowledge by feeding image tags and captions as the input prompt. They integrate implicit and explicit knowledge and generate the final answer by integrating both implicit and explicit knowledge sources. The authors of~\cite{yang2022empirical} propose PICa which relies on GPT-3 in joint retrieval of relevant knowledge and reasoning. They verbalize the visual signal by replacing the image with its caption and/or tags to provide more in-context information for the language model. 

\citeauthor{lin2022revive} offer REVIVE to integrate visual information, implicit knowledge, and explicit knowledge to generate the final answer. Similar to KAT, they have used a subset of Wikidata \cite{vrandevcic2014wikidata} as their KB. The authors of~\cite{hu2023promptcap} propose Promptcap (Prompt-guided image Captioning) to control the visual entities in the generated caption using a textual query and replace a general caption with a question-dependent caption. 

To avoid losing visual information, the authors of~\cite{salaberria2023image} feed visual features alongside regional labels and captions to the transformer in their proposed Caption-Based Model(CBM). Prophet's authors suggest adding candidate answers to enrich the prompt and improve final accuracy for LLM-based VQA~\cite{shao2023prompting}.

The target in implicit-KB VQA models is using LLMs as the source of knowledge, which needs verbalizing the images using a multimodal/image-captioning model. Therefore, the main effort of LLM-based models is to find the best verbalization for the images to avoid loss of required visual information.

\section{The Proposed Method}

Our goal is to solve the Knowledge-Based Visual Question Answering problem (KB-VQA). The input to the problem is an image and a question in natural language form. The goal is to find a short textual answer to the question. The answer to the question not only needs understanding the visual information but also requires external knowledge that is not contained in the image. In this problem setting in some cases, the source of external knowledge is also provided in the form of a knowledge graph~\cite{marino2021krisp}. In this paper, we do not exploit this explicit source of knowledge to find the answer and as we explain later we rely on the implicit knowledge of the LLMs in our proposed pipeline for finding the answer. 

\begin{figure} [htb] 
    \centering
    \includegraphics[width=1.0\linewidth]{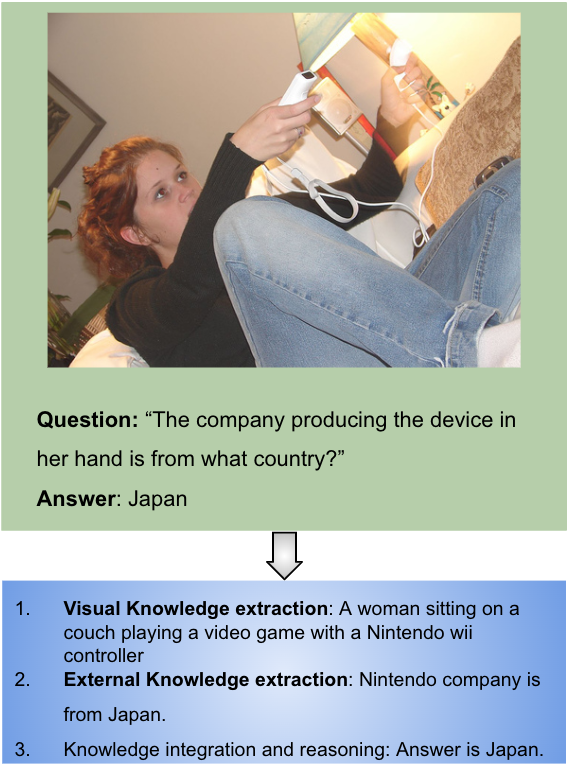}
    \caption{An example from the A-OKVQA dataset (green box). Given an image and a question, we must find the final answer by extracting and integrating visual and external knowledge (as shown in the blue box).}
    \label{fig:data}
\end{figure}

An example from the A-OKVQA dataset is shown in Fig~\ref{fig:data}. 

As shown in the blue box, answering this question requires visual knowledge (The device in her hand is a Nintendo Wii for playing games), some external knowledge beyond the given image (Nintendo Wii company is from Japan), and finally integrating all these sources of knowledge and reasoning over them to find the final answer (Japan).
 
\begin{figure} [htb] 
    \centering
    \includegraphics[width=0.8\linewidth]{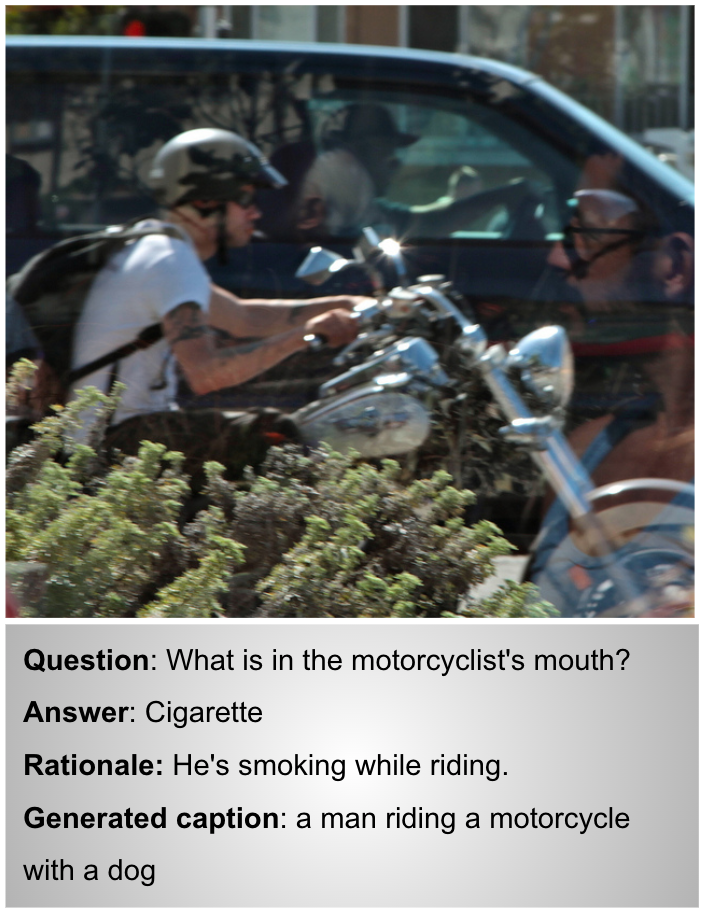}
    \caption{An example from the A-OKVQA dataset with their automatically generated captions. }
    \label{fig:cap3}
\end{figure}

Recently, most state-of-the-art VQA models use LLMs and their strength to answer visual questions, though these models trust captioners to verbalize the given image as shown in Fig~\ref{fig:cap3}.

\begin{figure*} [htb] 
    \centering
    \includegraphics[trim=0.2cm 0cm 0cm 0.5cm,clip, scale=0.30]{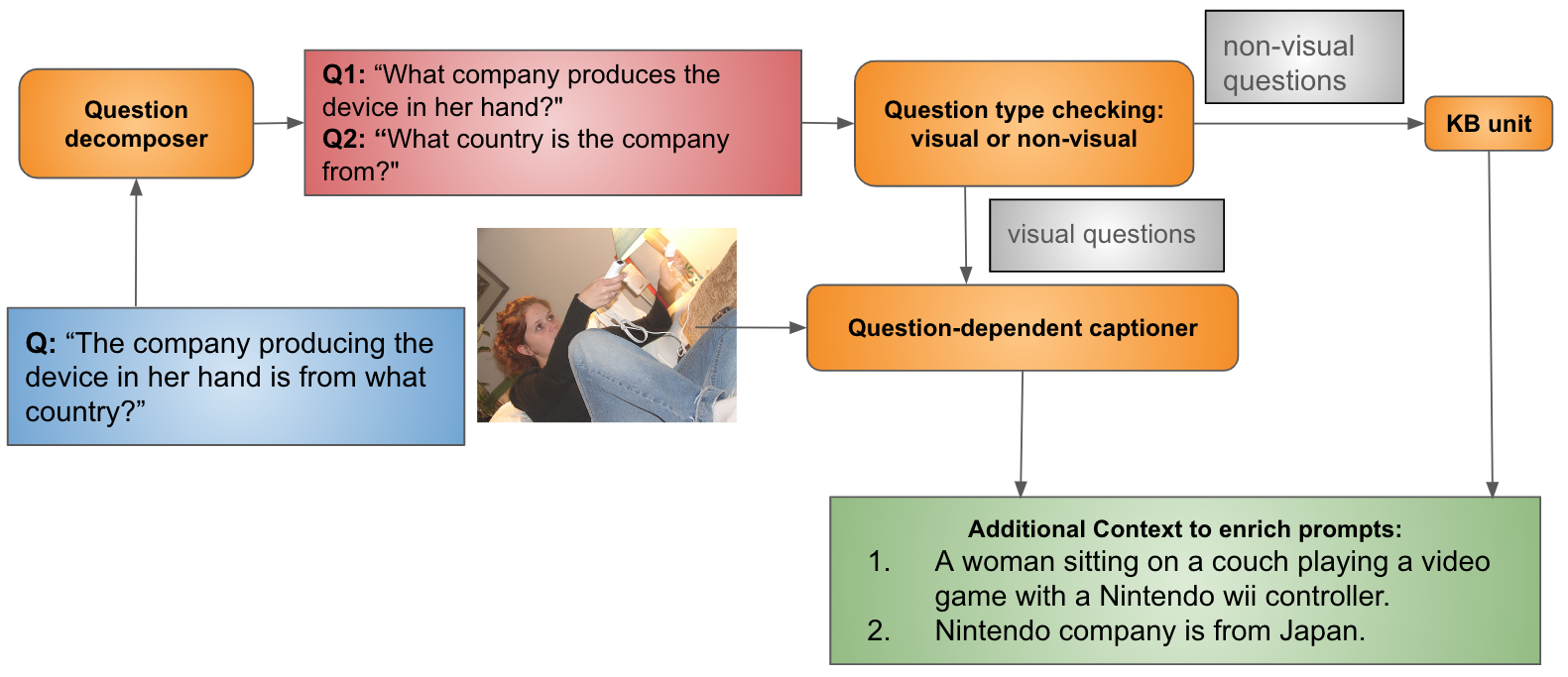}
    \caption{The Diagram of our model for question decomposition, type checking, information extraction, and final context collection.}
    \label{fig:arch}
\end{figure*}

Instead of passing a multi-hop or complicated question directly to the captioners, or using a general captioner, we aim to break it down into several simpler questions to prompt the captioner using various queries and extract all required information. In other words, our goal is to find a richer verbalization of the given image. It is worth mentioning that a "simpler" could be shorter, or any relevant similar question to help extract more visual information. An overview of our model is shown in Fig~\ref{fig:arch} and will be elaborated in the following sections.

We decompose each question using "gpt-3.5-turbo-0125" as follows:

\vspace{\parskip}
%\begin{minipage}{\textwidth}
\begin{tcolorbox}
Split this question into smaller ones. \\ "Question": "What is in the motorcyclist's mouth?" \\
       "Splits": "1. Is the motorcyclist wearing a helmet?\\
     2. Is the motorcyclist talking or eating something?\\
     3. Is the motorcyclist smoking or chewing gum?\\
     4. Can we see inside the motorcyclist's mouth?"

\end{tcolorbox}
%\end{minipage}
\vspace{\parskip}

Next, to find whether the question is visual or knowledge-based, we ask "gpt-3.5-turbo-0125" if the question needs external content to be answered or not as follows:

\vspace{\parskip}
\begin{tcolorbox}
Please answer with yes or no. Do you need external information (e.g. visual, verbal, contextual, ...) to answer this question?
\end{tcolorbox}
\vspace{\parskip}

If the answer to the above question is yes, the question is a visual question implying that we need to ask a captioner to provide a question-conditioned caption. In this case, we prompt our captioner to extract the relevant visual knowledge from the image. %If the captioner gives us a \pk{complete sentence: [what do you mean here?]}, like PromptCap, we do not need further steps with visual information extraction. 
For models like InstructBlip which provide a short answer to a given question, we take an additional step and use the following prompt for "gpt-3.5-turbo-0125" to rephrase the visual question and its generated answer and generate a complete sentence. We add the generated full sentence to the final context.
\vspace{\parskip}
\begin{tcolorbox}
Rephrase the question and answer into a single statement.
The re-phrased statement should summarize the question and answer.
The re-phrased statement should not be a question. 
\textbf{Question}: what is on top of the cupcakes? \\
\textbf{Answer}: white frosting and a cherry. \\
\textbf{Full-sentence generated by GPT}: The cupcakes are topped with white frosting and a cherry.
\end{tcolorbox}
\vspace{\parskip}

For the non-visual questions, we can use any knowledge base to extract extra non-visual knowledge. We have simply used "gpt-3.5-turbo-0125" with the following command.

\vspace{\parskip}
\begin{tcolorbox}
Please answer the following question:\\
Question: Whose responsibility is it to maintain the cleanliness of the toilet? \\
Answer: It is generally the responsibility of the owner or resident of the household to maintain the cleanliness of the toilet.
\end{tcolorbox}
\vspace{\parskip}

Finally, we use the above-extracted context to prompt the GPT model and generate the final answer. 
Similar to \cite{hu2023promptcap,shao2023prompting, yang2022empirical}, we use a combination of the question and image caption as the input prompt for our LLM building block. 
For each test example, we find 32 training examples using maximum cosine similarity of image+question embeddings to form our few-shot prompts. We use CLIP\cite{radford2021learning} to embed the questions and images and find the best few-shots for each text example.

\begin{figure*} [htb] 
    \centering
    \includegraphics[trim=0.1cm 0cm 0cm 0.3cm,clip, scale=0.28]{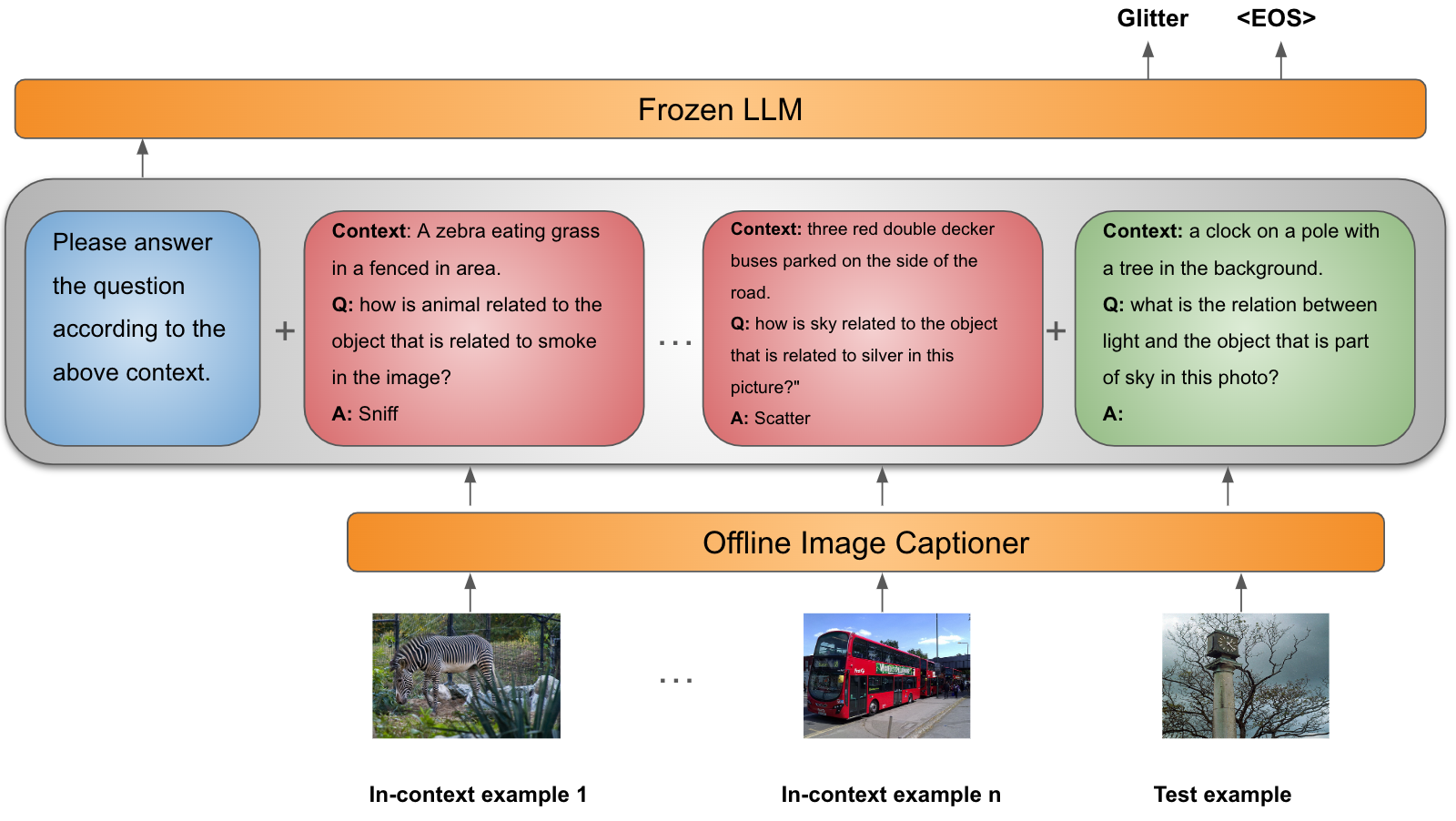}
    \caption{General architecture of using LLMs for VQA model. The context extracted by architecture in Fig~\ref{fig:arch} will be used in this few-shot LLM model to find the final answer.}
    \label{fig:LLM}
\end{figure*}

Due to its significant accuracy of "davinci-002" for in-context learning, like \cite{hu2023promptcap, shao2023prompting} we use it as our LLM engine as can be seen in Figure \ref{fig:LLM}.  

\subsection{Adding OCR to context}
\begin{figure} [htb] 
    \centering
    \includegraphics[width=0.99\linewidth]{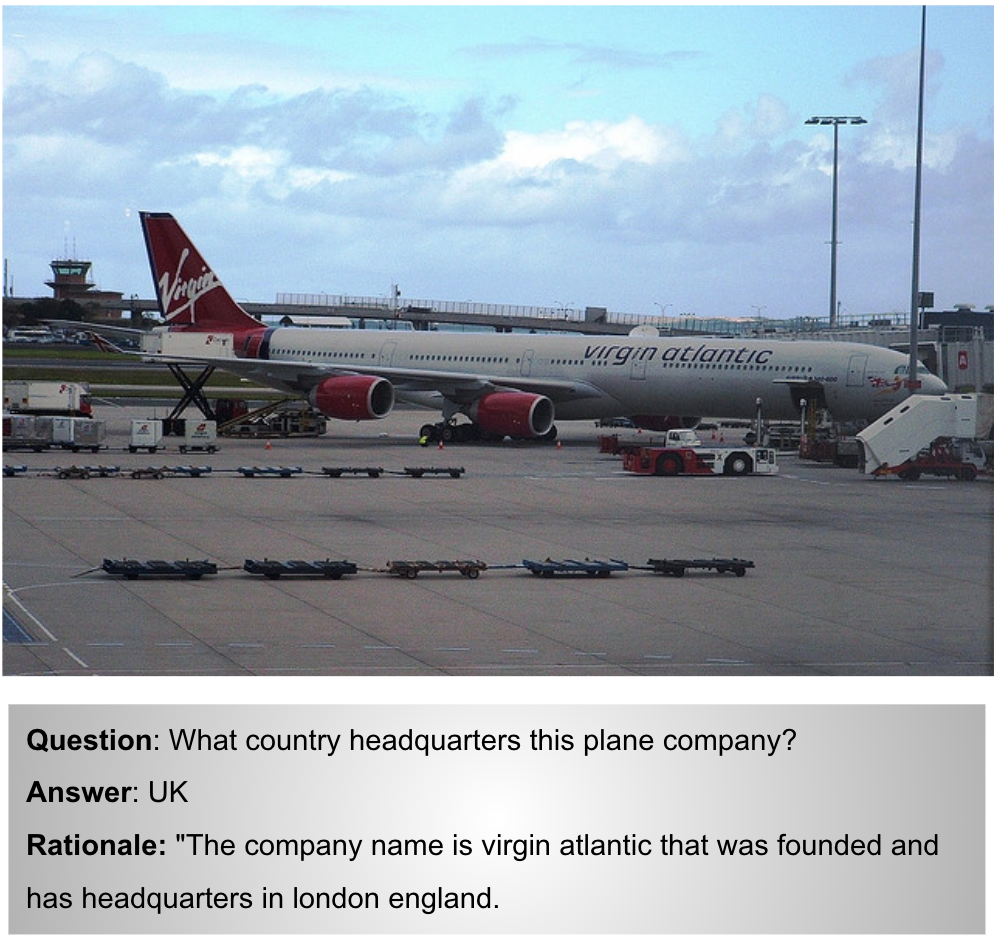}
    \caption{An example from A-OKVQA dataset with the image taken from COCO dataset\cite{1}}
    \label{fig:ocr1}
\end{figure}

To address the visual requirements of a question for a VQA task, we need to have a strong comprehension of an image that covers not only the main message of the image but also details such as written text in the image. We have shown one example from the A-OKVQA task in Fig.~\ref{fig:ocr1} which requires extracting written text in the image to find the final answer. Since most of the general captioners are not strong in perceiving OCR, we used a model called EasyOCR \cite{jaide2023} which extracts written text in  $[[c_1, c_2,c_3,c_4], "text", p]]$ format that includes the geometrical dimension of the box, the written text, and its corresponding probability. We have added OCR tokens for each image to the context in $"text" (p),$ format which includes tokens and their corresponding probability, and ignore bounding boxes, since it generates more stable results than the other combinations.

%\footnotetext{http://images.cocodataset.org/val2017/000000495054.jpg}

\section{Experimental Results}

In the following sections, we provide experimental results to show the impact of enriching visual and non-visual context by decomposing the original question into multiple simpler questions. 

\subsection{Datasets}\label{sec:dataset}

OK-VQA is a commonly used knowledge-based VQA dataset~\cite{marino2019ok}. The dataset contains 9K and 5K image-question pairs for training and testing, respectively. All questions are manually filtered to ensure that outside knowledge is required to answer the questions. Each data sample is annotated with ten open-ended answers. The accuracy computed by the soft scores is used as the evaluation metric~\cite{goyal2017making}.

A-OKVQA is a large knowledge-based VQA dataset~\cite{schwenk2022okvqa}. The dataset is split into three subsets: 17K training, 1K validation, and 7K testing. Each question is annotated with ten open-ended answers for direct answer (DA) evaluation. Besides, it provides a multiple choice (MC) evaluation to choose the correct answer from four choices. Since the validation set is supported with corresponding rationales, we have reported the final result on the validation set. 

KRVQA is to date the largest knowledge-based VQA dataset \cite{cao2021knowledge}. It contains 32,910 images and 157,201 QA pairs. They are split into the train, validation, and test set at the ratio of 60\%, 20\%, and 20\%, respectively.  The questions in this dataset use external knowledge from FVQA dataset 225434 factual triplets~\cite{wang2017fvqa}. %This external knowledge is being extracted from DBpedia\cite{auer2007dbpedia}, ConceptNet\cite{speer2017conceptnet}, and WebChild\cite{tandon2014webchild}. 
KRVQA authors aim to prevent shortcut learning by generating non-crowdsourced question-answer pairs with their supporting reasons indicating which part of the reason is from visual, and which part is from external knowledge.  %It provides meta-data to analyze the strengths and weaknesses of the VQA models, for example, question type, supporting reasons divided into KB and visual parts, and number of reasoning hops to answer the questions. 
\subsection{Implementation Details} \label{sec:imp}
 We implement our model with PyTorch \cite{paszke2019pytorch} for all experiments using one GPU with 49GB of memory.  We use top-1 accuracy for a fair comparison.
 
\subsection{Result and Discussion}\label{sec:acc}
We have reported accuracy for 3 of the recently popular VQA methods PICA, PromptCAP, and Prophet, and compared to our proposed method in Table \ref{tab:acc}. PICA \cite{yang2022empirical} uses GPT-3 as the main engine, and replaces the image using a general captioner to provide more in-context information for the language model. PromptCap~\citeauthor{hu2023promptcap} uses GPT3 as the main engine, though fine-tunes a general captioner to provide question-dependent captions instead of general captions. \citeauthor{shao2023prompting} in Prophet uses a vanilla VQA model, MCAN \cite{yu2019deep}, to provide 10 candidate answers to enrich the contextual information. In Prophet+PromptCap, we combined both PromptCap captioner~\citeauthor{hu2023promptcap} and 10 candidate answers to check the improvement made by adding both question-based captioner and injecting candidates in the context.
For a fair comparison, we have used the same version of the LLM engine in all models and reported their accuracy in an equivalent setting. 
\begin{table*}[htb!]
    \centering
    \small
    \scalebox{0.95}{
    \begin{tabular}{cccc} 
Models/Accuracy      & AOKVQA & OKVQA & KRVQA  \\ \toprule
PICA \cite{yang2022empirical}                 & 45.59 & 50.48 &  17.3 \\ \midrule
PromptCap \cite{hu2023promptcap}              & 47.66 & 55.40 & 19.6  \\ \midrule
Prophet \cite{shao2023prompting}              & 54.37 & 56.32 & 26.5  \\ \midrule
Prophet+PromptCap      & 54.65 & 56.29 & 26.2  \\ \midrule
Ours with PromptCap    & 56.2  & 56.71 & 27.4  \\ \midrule
Ours with InstructBlip & 56.77 & 57.05 & 27.8  \\ \bottomrule

\end{tabular}
    }
    \caption{Top-1 Accuracy for our proposed method compared to SOTA methods.}
    \label{tab:acc}
\end{table*}
It is worth mentioning that we used several different captioners for decomposed visual questions; among all PromptCap and IntructBlip provided the most stable results. Therefore, We reported the results of using PromptCap vs InstructBLIP as the visual unit for the simplified questions. In both cases, we achieve a 1 to 2\% improvement in accuracy, though InstructBlip works slightly better. After analyzing the test results, we conjecture that this is because Promptcap always aims to modify the general caption slightly to address the given question, while InstructBLIP has much more strength and freedom to address the details of the image to satisfy the given question. 

\subsection{Ablation Study}

We report the results of our ablation study in Table \ref{tab:ablation} to find the impact of adding different information to the context.  
Our BASE model uses GPT3 as the main engine, PromptCap as the question-dependent captioner, and Prophet adds top-10 candidates to the context of each example.
\begin{table*}[htb!]
\centering
\small
\renewcommand{\arraystretch}{1}
\scalebox{0.95}{
\begin{tabular}{cccc} 
Models/Accuracy & AOKVQA & OKVQA & KRVQA \\ \toprule
BASE       & 54.65 & 56.29 & 26.8 \\ \midrule
+OCR       & 54.83 & 57.43 & 26.9 \\ \midrule
+decomp with Promptcap as captioner    & 55.80 & 56.75 & 27.1 \\ \midrule
+decomp with InstructBlip as captioner &  \textbf{57.44} & 56.16 & 27.7 \\ \midrule
+decomp with Promptcap as captioner+OCR      &  56.2 & 56.71 & 27.4 \\ \midrule
+decomp with InstructBlip as captioner+OCR & 56.77 & \textbf{57.05} & \textbf{27.8} \\ \bottomrule
 \end{tabular}
    }
    \caption{Ablation study of our model.}
    \label{tab:ablation}
\end{table*}
We investigate using PromptCap as the visual unit for the simplified decomposed questions vs InstructBlip. InstructBlip works superior to PromptCap always. Following, we show an example with the new questions generated by our decomposition unit.  By comparing the OCR vs decomposition gain for OKVQA, we find that OCR brings more improvement than decomposition for OKVQA dataset which shows OKVQA is an older and simpler dataset than A-OKVQA and KRVQA which have been designed to evaluate reasoning ability and handling multiple hops in the question.

An example of our decomposition model is shown below. Our model has generated 4 questions to help extract relevant information from the image. We can see that questions aim to uncover "what is in the rider's mouth?" using various forms of questions regarding eating, smoking, moth visibility, and head cover.

\vspace{\parskip}
\begin{tcolorbox}
"question": "What is in the motorcyclist's mouth?", \\
       "splits": \\
     1. Is the motorcyclist wearing a helmet?\\
     2. Is the motorcyclist talking or eating something?\\
     3. Is the motorcyclist smoking or chewing gum?\\
     4. Can we see inside the motorcyclist's mouth?
\end{tcolorbox}
\vspace{\parskip}

The PromptCap model generates the following content for each of the above splits. It is clear from these answers that PromptCap often repeats the salient objects of the given image with slight modifications toward the given question, if any. Moreover,  sometimes these modifications result in wrong information, such as the 4th question.
\vspace{\parskip}
\begin{tcolorbox}
           "1. a man wearing a helmet riding a motorcycle with a dog", \\
           "2. a man riding a motorcycle with a dog", \\
           "3. a man riding a motorcycle with a dog", \\
           "4. a man riding a motorcycle with a dog in his mouth"
\end{tcolorbox}
\vspace{\parskip}

The visual knowledge extracted through InstructBlip is as in the example in the following box. We can see that InstructBlip has a better perception of the image and does not make up the content if it cannot extract the relevant details.
\vspace{\parskip}
\begin{tcolorbox}
1. The motorcyclist is wearing a helmet., \\
2. The motorcyclist is talking., \\
3. The motorcyclist is smoking., \\
4. We cannot see inside the motorcyclist's mouth, 
\end{tcolorbox}
\vspace{\parskip}

We can see that adding OCR tokens to the BASE model results in slight improvements for the A-OKVQA and KRVQA datasets. More comparative improvements are observed for the OKVQA dataset. This result demonstrates the existence of many questions requiring written text extraction from the images in the OKVQA dataset. Our simple method of adding OCR information to the context improves the final accuracy most often since it adds some knowledge ignored by the general captioners, though this improvement is not consistent across all settings which can be due to some context conflicts. 

\begin{figure} [htb] 
    \centering
    \includegraphics[width=0.85\linewidth]{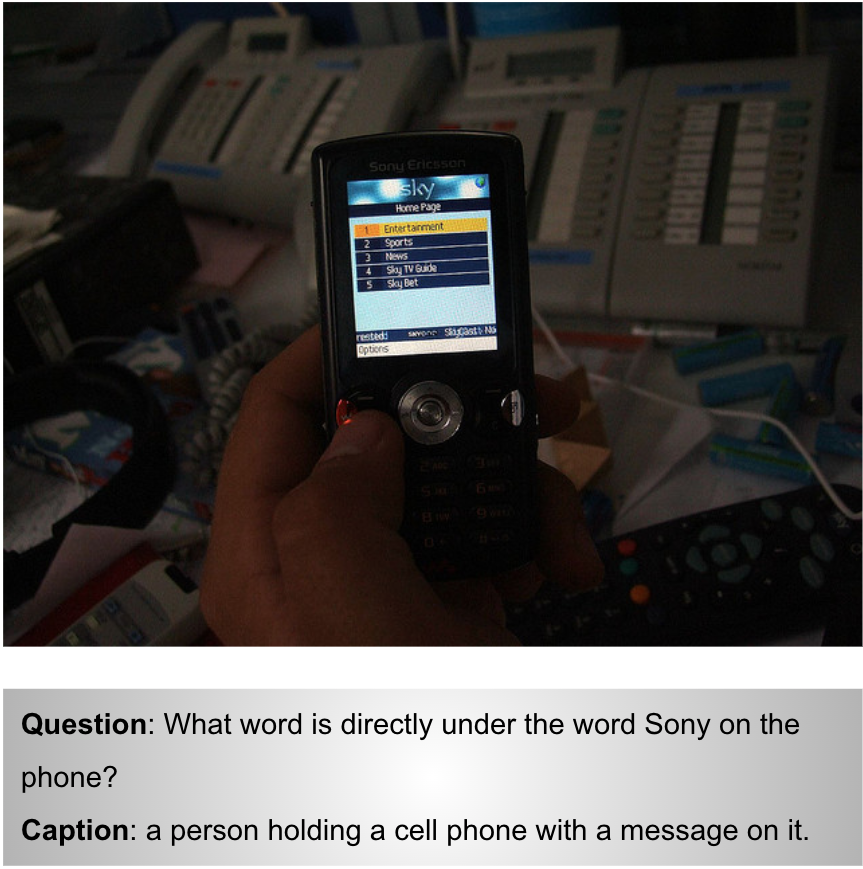}
\caption{A question-image pair from A-OKVQA dataset with automatically generated caption\cite{2}.}
\label{fig:OCR}
\end{figure}

In the future, we train/fine-tune a strong captioner to perceive and explain the written text. An example is shown in Fig \ref{fig:OCR} which not only requires strong written text extraction, but their relative geometrical information as well. Therefore, training a strong visual element to perceive and explain OCR content is vital for VQA tasks.

\begin{comment}
\begin{table*}[h!]
\centering
\small
\scalebox{0.99}{
\begin{tabular}{cccc} 
Models/Accuracy & AOKVQA & OKVQA & KRVQA \\ \toprule
BASE & 47.66 & 55.40 & 19.6  \\ \midrule
+ OCR & 48.31 & 55.28 & 18.7 \\ \midrule
+ 10 candidates       & 54.65 & 56.29 & 26.8 \\ \midrule
+ OCR + 10 candidates & 54.83 & 57.43 & 26.9 \\ \midrule
+decomp & 49.44 & 56.31 & 17.4 \\ \midrule
+decomp+OCR & 47.42 & 52.61 & 17.5 \\ \midrule
+decomp+10cndd & 53.87 & 56.72 & 27.6 \\ \midrule
+decomp with Promptcap as captioner+OCR+10cndds      &  56.2 & 56.71 & 27.4 \\ \midrule
+decomp with instructblip as captioner+OCR+10cndds & 56.77 & 57.05 & 27.8 \\ \bottomrule
 \end{tabular}
    }
    \caption{Ablation study of our model.}
    \label{tab:ablation}
\end{table*}
\end{comment}
\section{Conclusion and Future Work}
Most of the current well-known B-VQA models rely on the power of LLMs for answer prediction and replace the given image with its textual version (caption). However, the captioning models are not strong enough to extract the required details from the image and only describe the main message of the image. Moreover, captioners either do not get the question or cannot understand and process complicated and multi-hop questions to generate a relevant caption. We propose to decompose the given question into several simpler and shorter questions and run a question-dependent captioner to provide a richer verbal translation of the given image which addresses the question requirements. Moreover, we enrich the visual context by extracting the OCR information from the images. In addition, by dividing the decomposed questions into visual and non-visual, we use the non-visual decomposed questions to extract information from other knowledge sources.
Our results demonstrate the positive impact of decomposing the original questions and generating several detailed captions using these questions. It is necessary to train a strong captioner who can comprehend images from different aspects including extracting written texts and their relevant information. 
\section{Limitations}

To enrich the visual context, we extracted and added the OCR tokens and their corresponding probability to the context while training a strong captioner to find OCR-related captions is vital for the VQA task.
    
Since the decomposed questions are not independent, a strong chat-based captioner is needed for type checking and information extraction for decomposed questions. The current chat-based captioners are not strong enough in comparison to the other captioners like InstructBlip and PromptCap.  Our efforts with chatBlip degraded the results, despite having a better understanding of the decomposed questions. A chat-based captioner can improve both type-checking and information extraction. 

Moreover, We used LLMs as an extra source of knowledge, though using other explicit sources of knowledge can make more improvement. In the future, we will try the same proposed method joined with explicit knowledge retrieval.

%\subsection{Acknowledgement}
%This project is supported by the National Science Foundation (NSF) CAREER award 2028626 and partially supported by the Office of Naval Research (ONR) grant N00014-20-1-2005 and grant N00014-23-1-2417. Any opinions, findings, and conclusions or recommendations expressed in this material are those of the authors and do not necessarily reflect the views of the National Science Foundation nor the Office of Naval Research. We thank all reviewers for their thoughtful comments and suggestions.
%\subsection{Appendix}
\bibliographystyle{acl_natbib}
\bibliography{anthology}

\begin{thebibliography}{31}
\expandafter\ifx\csname natexlab\endcsname\relax\def\natexlab#1{#1}\fi

\bibitem[{jai()}]{jaide2023}

\newblock \href {https://jaided.ai/easyocr/install/} {Easyocr}.
\newblock \url{https://jaided.ai/easyocr/install/}.
\newblock Online; accessed 12 June 2024.

\bibitem[{1()}]{1}

\newblock \href {http://images.cocodataset.org/train2017/000000286328.jpg} {Mscoco dataset 1}.
\newblock Online; accessed 12 June 2024.

\bibitem[{2()}]{2}

\newblock \href {http://images.cocodataset.org/val2017/000000495054.jpg} {Mscoco dataset 2}.
\newblock Online; accessed 12 June 2024.

\bibitem[{Ben-Younes et~al.(2017)Ben-Younes, Cadene, Cord, and Thome}]{ben2017mutan}
Hedi Ben-Younes, R{\'e}mi Cadene, Matthieu Cord, and Nicolas Thome. 2017.
\newblock Mutan: Multimodal tucker fusion for visual question answering.
\newblock In \emph{Proceedings of the IEEE international conference on computer vision}, pages 2612--2620.

\bibitem[{Brown et~al.(2020)Brown, Mann, Ryder, Subbiah, Kaplan, Dhariwal, Neelakantan, Shyam, Sastry, Askell et~al.}]{brown2020language}
Tom Brown, Benjamin Mann, Nick Ryder, Melanie Subbiah, Jared~D Kaplan, Prafulla Dhariwal, Arvind Neelakantan, Pranav Shyam, Girish Sastry, Amanda Askell, et~al. 2020.
\newblock Language models are few-shot learners.
\newblock \emph{Advances in neural information processing systems}, 33:1877--1901.

\bibitem[{Cao et~al.(2021)Cao, Li, Liang, Wang, and Lin}]{cao2021knowledge}
Qingxing Cao, Bailin Li, Xiaodan Liang, Keze Wang, and Liang Lin. 2021.
\newblock Knowledge-routed visual question reasoning: Challenges for deep representation embedding.
\newblock \emph{IEEE Transactions on Neural Networks and Learning Systems}, 33(7):2758--2767.

\bibitem[{Dai et~al.(2024)Dai, Li, Li, Tiong, Zhao, Wang, Li, Fung, and Hoi}]{dai2024instructblip}
Wenliang Dai, Junnan Li, Dongxu Li, Anthony Meng~Huat Tiong, Junqi Zhao, Weisheng Wang, Boyang Li, Pascale~N Fung, and Steven Hoi. 2024.
\newblock Instructblip: Towards general-purpose vision-language models with instruction tuning.
\newblock \emph{Advances in Neural Information Processing Systems}, 36.

\bibitem[{Ding et~al.(2022)Ding, Yu, Liu, Hu, Cui, and Wu}]{ding2022mukea}
Yang Ding, Jing Yu, Bang Liu, Yue Hu, Mingxin Cui, and Qi~Wu. 2022.
\newblock Mukea: Multimodal knowledge extraction and accumulation for knowledge-based visual question answering.
\newblock In \emph{Proceedings of the IEEE/CVF Conference on Computer Vision and Pattern Recognition}, pages 5089--5098.

\bibitem[{Goyal et~al.(2017)Goyal, Khot, Summers-Stay, Batra, and Parikh}]{goyal2017making}
Yash Goyal, Tejas Khot, Douglas Summers-Stay, Dhruv Batra, and Devi Parikh. 2017.
\newblock Making the v in vqa matter: Elevating the role of image understanding in visual question answering.
\newblock In \emph{Proceedings of the IEEE conference on computer vision and pattern recognition}, pages 6904--6913.

\bibitem[{Gui et~al.(2021)Gui, Wang, Huang, Hauptmann, Bisk, and Gao}]{gui2021kat}
Liangke Gui, Borui Wang, Qiuyuan Huang, Alex Hauptmann, Yonatan Bisk, and Jianfeng Gao. 2021.
\newblock Kat: A knowledge augmented transformer for vision-and-language.
\newblock \emph{arXiv preprint arXiv:2112.08614}.

\bibitem[{Guo et~al.(2022)Guo, Nie, Wong, Liu, Cheng, and Kankanhalli}]{guo2022unified}
Yangyang Guo, Liqiang Nie, Yongkang Wong, Yibing Liu, Zhiyong Cheng, and Mohan Kankanhalli. 2022.
\newblock A unified end-to-end retriever-reader framework for knowledge-based vqa.
\newblock In \emph{Proceedings of the 30th ACM International Conference on Multimedia}, pages 2061--2069.

\bibitem[{Hu et~al.(2023)Hu, Hua, Yang, Shi, Smith, and Luo}]{hu2023promptcap}
Yushi Hu, Hang Hua, Zhengyuan Yang, Weijia Shi, Noah~A Smith, and Jiebo Luo. 2023.
\newblock Promptcap: Prompt-guided image captioning for vqa with gpt-3.
\newblock In \emph{Proceedings of the IEEE/CVF International Conference on Computer Vision}, pages 2963--2975.

\bibitem[{Khare et~al.(2021)Khare, Priyakumar, BAGAL, Jawahar, Devi, and Mathew}]{khare2021mmbert}
Yash Khare, U~Priyakumar, VIRAJ BAGAL, CV~Jawahar, Adithi Devi, and Minesh Mathew. 2021.
\newblock Mmbert: Multimodal bert pretraining for improved medical vqa.

\bibitem[{Li and Moens(2022)}]{li2022dynamic}
Mingxiao Li and Marie-Francine Moens. 2022.
\newblock Dynamic key-value memory enhanced multi-step graph reasoning for knowledge-based visual question answering.
\newblock In \emph{Proceedings of the AAAI Conference on Artificial Intelligence}, volume~36, pages 10983--10992.

\bibitem[{Lin et~al.(2022)Lin, Xie, Chen, Xu, Zhu, and Yuan}]{lin2022revive}
Yuanze Lin, Yujia Xie, Dongdong Chen, Yichong Xu, Chenguang Zhu, and Lu~Yuan. 2022.
\newblock Revive: Regional visual representation matters in knowledge-based visual question answering.
\newblock \emph{arXiv preprint arXiv:2206.01201}.

\bibitem[{Marino et~al.(2021)Marino, Chen, Parikh, Gupta, and Rohrbach}]{marino2021krisp}
Kenneth Marino, Xinlei Chen, Devi Parikh, Abhinav Gupta, and Marcus Rohrbach. 2021.
\newblock Krisp: Integrating implicit and symbolic knowledge for open-domain knowledge-based vqa.
\newblock In \emph{Proceedings of the IEEE/CVF Conference on Computer Vision and Pattern Recognition}, pages 14111--14121.

\bibitem[{Marino et~al.(2019)Marino, Rastegari, Farhadi, and Mottaghi}]{marino2019ok}
Kenneth Marino, Mohammad Rastegari, Ali Farhadi, and Roozbeh Mottaghi. 2019.
\newblock Ok-vqa: A visual question answering benchmark requiring external knowledge.
\newblock In \emph{Proceedings of the IEEE/cvf conference on computer vision and pattern recognition}, pages 3195--3204.

\bibitem[{Paszke et~al.(2019)Paszke, Gross, Massa, Lerer, Bradbury, Chanan, Killeen, Lin, Gimelshein, Antiga et~al.}]{paszke2019pytorch}
Adam Paszke, Sam Gross, Francisco Massa, Adam Lerer, James Bradbury, Gregory Chanan, Trevor Killeen, Zeming Lin, Natalia Gimelshein, Luca Antiga, et~al. 2019.
\newblock Pytorch: An imperative style, high-performance deep learning library.
\newblock \emph{Advances in neural information processing systems}, 32.

\bibitem[{Radford et~al.(2021)Radford, Kim, Hallacy, Ramesh, Goh, Agarwal, Sastry, Askell, Mishkin, Clark et~al.}]{radford2021learning}
Alec Radford, Jong~Wook Kim, Chris Hallacy, Aditya Ramesh, Gabriel Goh, Sandhini Agarwal, Girish Sastry, Amanda Askell, Pamela Mishkin, Jack Clark, et~al. 2021.
\newblock Learning transferable visual models from natural language supervision.
\newblock In \emph{International conference on machine learning}, pages 8748--8763. PMLR.

\bibitem[{Salaberria et~al.(2023)Salaberria, Azkune, de~Lacalle, Soroa, and Agirre}]{salaberria2023image}
Ander Salaberria, Gorka Azkune, Oier~Lopez de~Lacalle, Aitor Soroa, and Eneko Agirre. 2023.
\newblock Image captioning for effective use of language models in knowledge-based visual question answering.
\newblock \emph{Expert Systems with Applications}, 212:118669.

\bibitem[{Schlichtkrull et~al.(2018)Schlichtkrull, Kipf, Bloem, Van Den~Berg, Titov, and Welling}]{schlichtkrull2018modeling}
Michael Schlichtkrull, Thomas~N Kipf, Peter Bloem, Rianne Van Den~Berg, Ivan Titov, and Max Welling. 2018.
\newblock Modeling relational data with graph convolutional networks.
\newblock In \emph{The Semantic Web: 15th International Conference, ESWC 2018, Heraklion, Crete, Greece, June 3--7, 2018, Proceedings 15}, pages 593--607. Springer.

\bibitem[{Schwenk et~al.(2022)Schwenk, Khandelwal, Clark, Marino, and Mottaghi}]{schwenk2022okvqa}
Dustin Schwenk, Apoorv Khandelwal, Christopher Clark, Kenneth Marino, and Roozbeh Mottaghi. 2022.
\newblock A-okvqa: A benchmark for visual question answering using world knowledge.
\newblock In \emph{European Conference on Computer Vision}, pages 146--162. Springer.

\bibitem[{Shao et~al.(2023)Shao, Yu, Wang, and Yu}]{shao2023prompting}
Zhenwei Shao, Zhou Yu, Meng Wang, and Jun Yu. 2023.
\newblock Prompting large language models with answer heuristics for knowledge-based visual question answering.
\newblock In \emph{Proceedings of the IEEE/CVF Conference on Computer Vision and Pattern Recognition}, pages 14974--14983.

\bibitem[{Tucker(1966)}]{tucker1966some}
Ledyard~R Tucker. 1966.
\newblock Some mathematical notes on three-mode factor analysis.
\newblock \emph{Psychometrika}, 31(3):279--311.

\bibitem[{Vrande{\v{c}}i{\'c} and Kr{\"o}tzsch(2014)}]{vrandevcic2014wikidata}
Denny Vrande{\v{c}}i{\'c} and Markus Kr{\"o}tzsch. 2014.
\newblock Wikidata: a free collaborative knowledgebase.
\newblock \emph{Communications of the ACM}, 57(10):78--85.

\bibitem[{Wang et~al.(2017)Wang, Wu, Shen, Dick, and Van Den~Hengel}]{wang2017fvqa}
Peng Wang, Qi~Wu, Chunhua Shen, Anthony Dick, and Anton Van Den~Hengel. 2017.
\newblock Fvqa: Fact-based visual question answering.
\newblock \emph{IEEE transactions on pattern analysis and machine intelligence}, 40(10):2413--2427.

\bibitem[{Wu et~al.(2022)Wu, Lu, Sabharwal, and Mottaghi}]{wu2022multi}
Jialin Wu, Jiasen Lu, Ashish Sabharwal, and Roozbeh Mottaghi. 2022.
\newblock Multi-modal answer validation for knowledge-based vqa.
\newblock In \emph{Proceedings of the AAAI Conference on Artificial Intelligence}, volume~36, pages 2712--2721.

\bibitem[{Yang et~al.(2022)Yang, Gan, Wang, Hu, Lu, Liu, and Wang}]{yang2022empirical}
Zhengyuan Yang, Zhe Gan, Jianfeng Wang, Xiaowei Hu, Yumao Lu, Zicheng Liu, and Lijuan Wang. 2022.
\newblock An empirical study of gpt-3 for few-shot knowledge-based vqa.
\newblock In \emph{Proceedings of the AAAI Conference on Artificial Intelligence}, volume~36, pages 3081--3089.

\bibitem[{Yu et~al.(2019)Yu, Yu, Cui, Tao, and Tian}]{yu2019deep}
Zhou Yu, Jun Yu, Yuhao Cui, Dacheng Tao, and Qi~Tian. 2019.
\newblock Deep modular co-attention networks for visual question answering.
\newblock In \emph{Proceedings of the IEEE/CVF conference on computer vision and pattern recognition}, pages 6281--6290.

\bibitem[{Yu et~al.(2018)Yu, Yu, Xiang, Fan, and Tao}]{yu2018beyond}
Zhou Yu, Jun Yu, Chenchao Xiang, Jianping Fan, and Dacheng Tao. 2018.
\newblock Beyond bilinear: Generalized multimodal factorized high-order pooling for visual question answering.
\newblock \emph{IEEE transactions on neural networks and learning systems}, 29(12):5947--5959.

\bibitem[{Zhu et~al.(2020)Zhu, Yu, Wang, Sun, Hu, and Wu}]{zhu2020mucko}
Zihao Zhu, Jing Yu, Yujing Wang, Yajing Sun, Yue Hu, and Qi~Wu. 2020.
\newblock Mucko: multi-layer cross-modal knowledge reasoning for fact-based visual question answering.
\newblock \emph{arXiv preprint arXiv:2006.09073}.

\end{thebibliography}

\end{document}